\documentclass[letterpaper, 10 pt, conference]{ieeeconf}  
\usepackage{graphicx}
\graphicspath{{figures/}} 
\usepackage{caption}
\usepackage{subcaption}
\usepackage{url}
\usepackage{multirow}
\usepackage{tikz}

\IEEEoverridecommandlockouts                              

\title{\LARGE \bf Acoustic Soft Tactile Skin (AST Skin)}

\author{Vishnu Rajendran S$^{1}$, Willow Mandil$^{1}$, Kiyanoush Nazari$^{1}$, Simon Parsons$^{1}$ and Amir Ghalamzan E.$^{2}$
\thanks{$^{1}$ University of Lincoln, Lincoln, UK $^{2}$ University of Surrey, Guildford, UK.}
}

\begin{document}

\maketitle
\thispagestyle{empty}
\pagestyle{empty}


\begin{abstract}
This paper presents a novel acoustic soft tactile (ATS) skin technology operating with sound waves. In this innovative approach, the sound waves generated by a speaker travel in channels embedded in a soft membrane and get modulated due to a deformation of the channel when pressed by an external force and received by a microphone at the end of the channel. The sensor leverages \emph{regression} and \emph{classification} methods for estimating the normal force and its contact location. Our sensor can be affixed to any robot part, e.g., end effectors or arm. We tested several regression and classifier methods to learn the relation between sound wave modulation, the applied force, and its location, respectively and picked the best-performing models for force and location predictions. Our novel tactile sensor yields 93\% of the force estimation within ±1.5 N tolerances for a range of 0-30$^{+1}$ N and estimates contact locations with over 96\% accuracy. We also demonstrated the performance of AST technology for a real-time gripping force control application.
\end{abstract}



\section{Introduction}
Despite advancements in soft tactile sensing, persistent limitations are evident (as discussed in a recent review on tactile sensing technologies~\cite{mandil2023tactile}). Many commercially available soft tactile sensors come with fixed shapes and sizes, posing challenges in integration with existing hardware, especially in space-constrained environments. For instance, incorporating camera-based tactile sensors is problematic due to their space requirements~\cite{yuan2017gelsight,lepora2021soft}. Furthermore, developing sensors with intricate electrical components to fit tight spaces demands sophisticated manufacturing techniques and incurs high costs. There is a need for less complex tactile sensing technologies that can easily adapt to various shapes, limited space and form factors. 

Accurate measurement of physical interactions plays a vital role in various physical robotic tasks, such as human-robot interaction~\cite{fritzsche2011tactile}, object grasping, and manipulation~\cite{yousef2011tactile}. Tactile sensing can be used to observe the interaction state in physical robot interactions\cite{nazari2021tactile}. Tactile predictive models~\cite{Mandil2022RSS} can then make predictive controller possible, e.g. for slippage control~\cite{nazari2023proactive} or cluster manipulation~\cite{nazari2023deep}, leading to safer, more precise, and more efficient actions in a broader range of physical interaction tasks~\cite{Mandil2022RSS,dahiya2009tactile,dargahi2005advances}. Soft tactile sensors are important for handling deformable objects and robust manipulation. These sensors feature a soft, flexible sensing surface whose deformation provides tactile information such as \emph{normal/shear forces}, \emph{contact location}, and contact surface texture, etc. Soft tactile sensors utilise both electronic (e.g., resistive~\cite{zimmer2019predicting}, capacitive~\cite{li2016wide}, piezoelectric~\cite{song2019pneumatic}, magnetic~\cite{rehan2022soft,diguet2022tactile}, impedance~\cite{wu2022new}) and non-electronic transduction methods (e.g., camera-based~\cite{ward2018tactip,gomes2022geltip,yuan2017gelsight,donlon2018gelslim,lambeta2020digit,10017344,sferrazza2019design}, fluid-based~\cite{gong2017pneumatic}, and acoustics~\cite{chuang2018ultrasonic,shinoda1997acoustic,tanaka2015tactile,teramoto2001acoustical}) and their combinations~\cite{park2022biomimetic} for converting the membrane deformation to tactile information.

\begin{figure*}[tb!]
    \centering
    \subfloat[]{\includegraphics[height=0.3\textwidth]{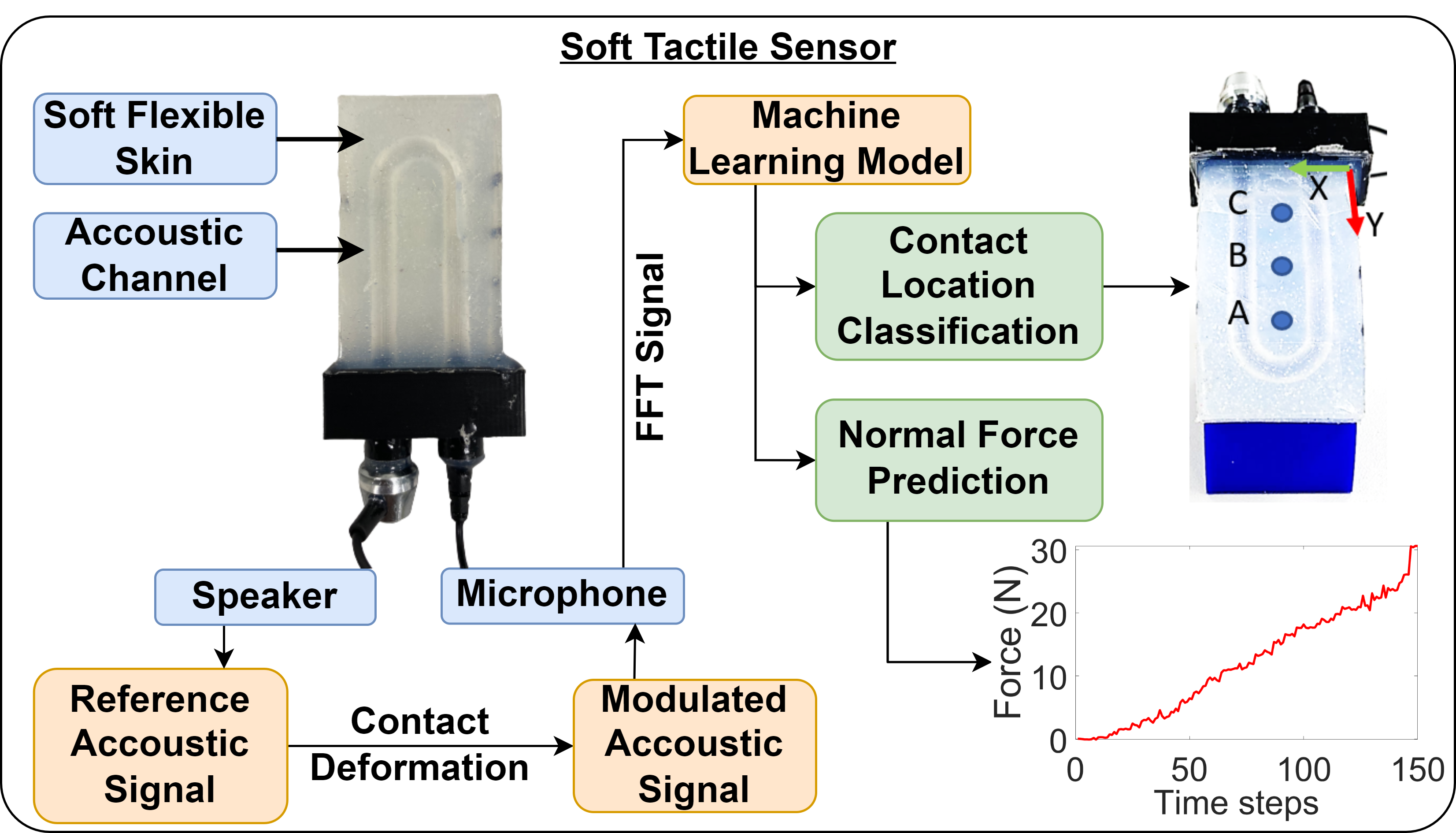}}
        \hspace{0.05cm}
    \subfloat[]{\includegraphics[height=0.3\textwidth]{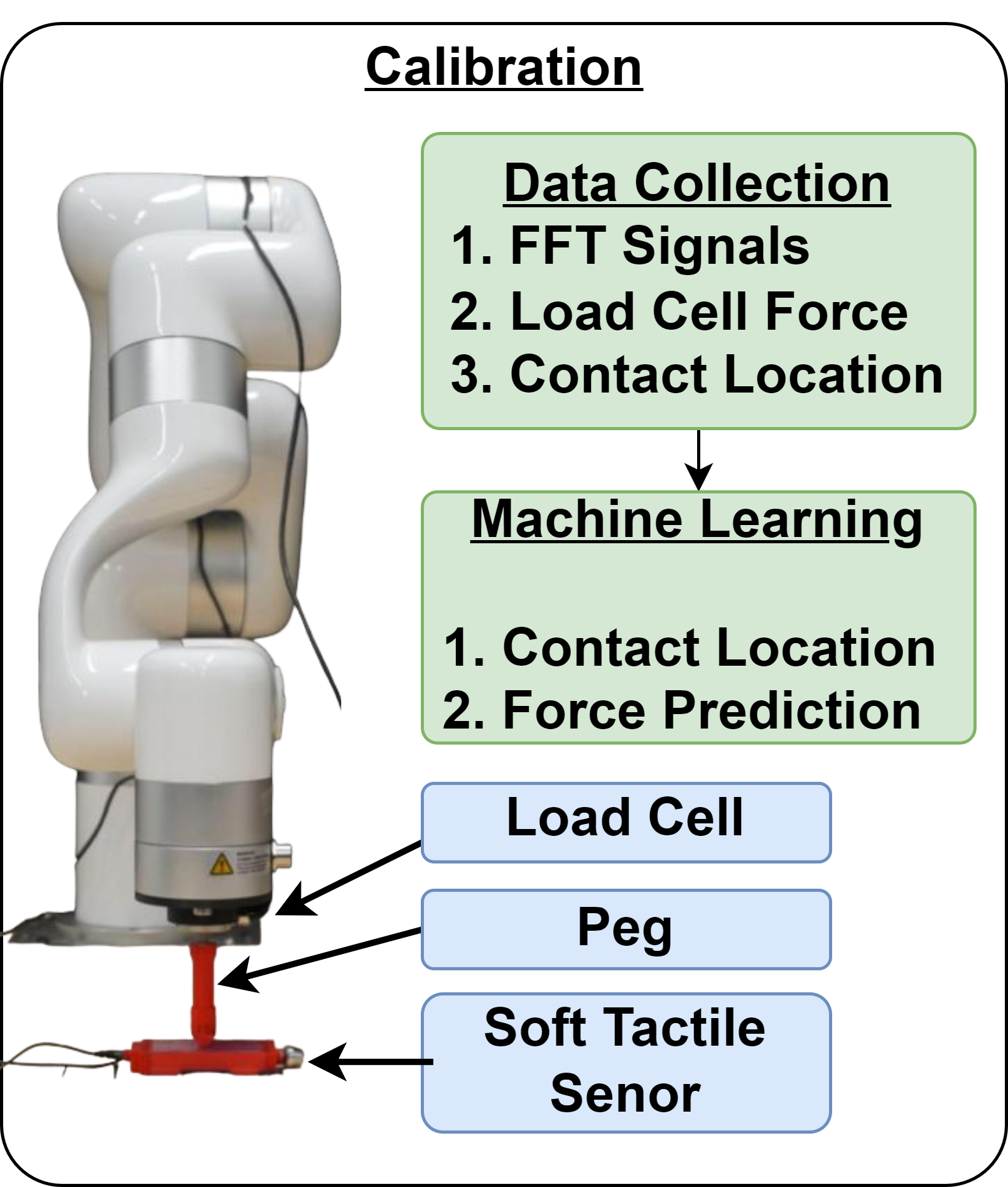}}
        \hspace{0.05cm}
    \subfloat[]{\includegraphics[height=0.3\textwidth]{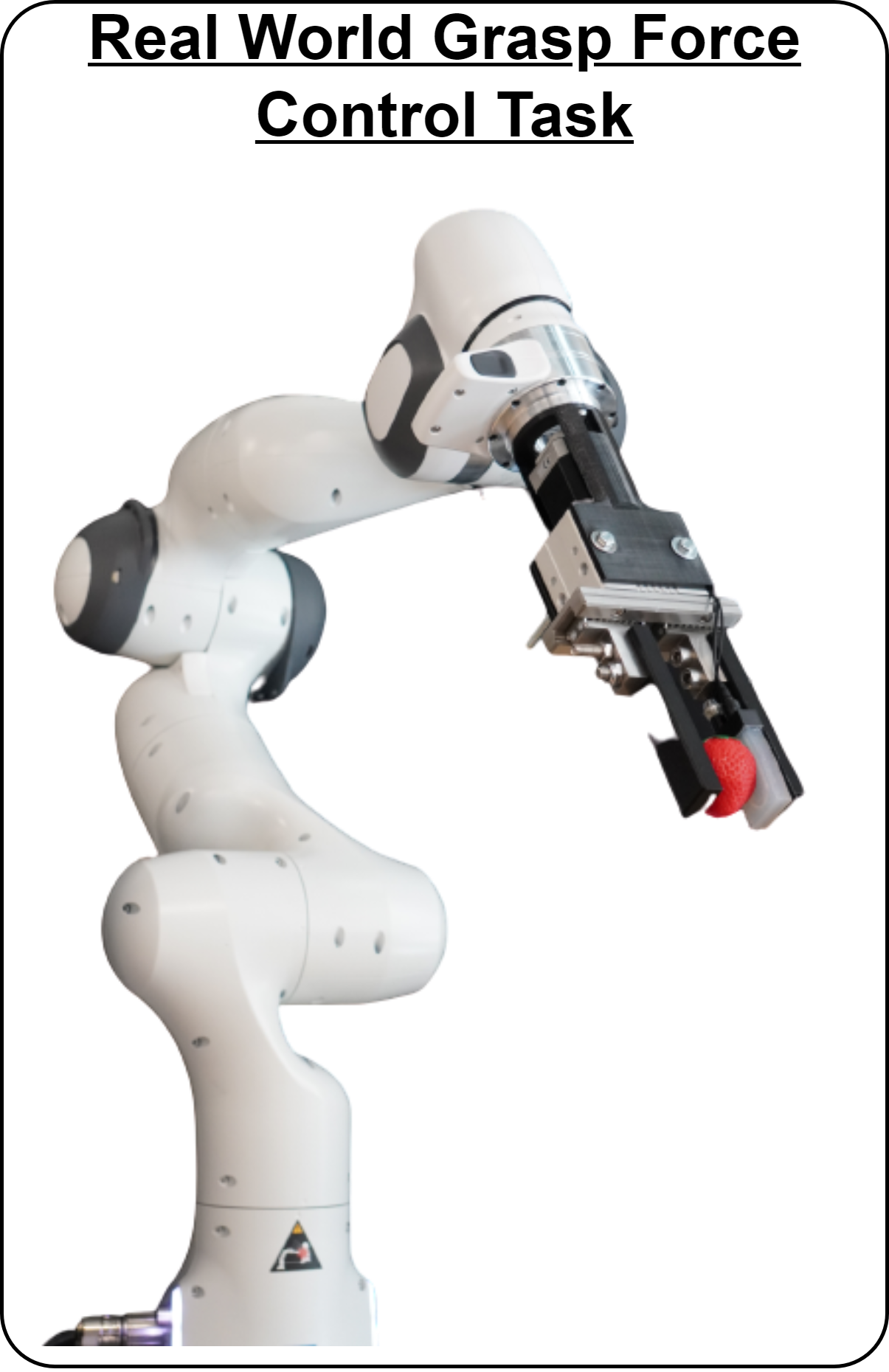}}
    \caption{AST Overview (a): A force applied to the AST surface deforms the acoustic channels beneath the sensing surface. These channels contain reference sound waves that travel from the speaker to the microphone. The sound wave amplitude is modulated in proportion to the deformation. We use FFT to transform the modulated waves to a frequency domain and machine learning methods to find the correlation between the frequencies and (i) the contact normal force and (ii) the contact location resulting from the deformation. (b) Sensor calibration: the xARM robot is used to apply a force to the AST at a set location through a load cell with a peg attached to the robot's wrist. The dataset contains the FFT signal, the load cell force, and the contact location. ML models are trained with this dataset to predict the contact force and location. (c) We apply this novel sensor to a grip force control pick-and-place task where the system grasps objects from the YCB object set \cite{calli2015benchmarking} (a soft strawberry in this example), adapts grip force to reach a target force, and then moves the object to a target location. We use the Franka Emika for this task and an SMC LEZH gripper.}
    \label{fig:setup}
    \vspace{-.4cm}
\end{figure*}

Zoller et al.~\cite{zoller2020active} proposed a simple concept for tactile sensing by using sound waves to a pneumatically actuated soft finger. They added a speaker and microphone within the finger's chamber for pneumatic actuation. The microphone continuously monitors changes in sound modulation. Sound waves' amplitude changes if the finger deforms due to an applied external force. They used the finger with added sensing capability to measure contact forces and contact location with a certain precision~\cite{wall2022passive}. However, the sensor is integrated with a soft finger and is not a stand-alone sensor to be used on other robot counterparts.

Acoustic techniques have been explored for deriving tactile features like contacts, deformations, forces, and shape recognition.
For instance, Shinoda et al. \cite{shinoda1996tactile} created a silicone hemispherical fingertip embedded with an ultrasound transmitter and receiver array, capable of detecting deformations up to 10µm and inclination changes to 0.001 radians. They later improved this design by adding a hollow cavity within the membrane to measure stress via acoustic resonance. Teramoto et al. \cite{teramoto2001acoustical} introduced a flexible membrane underlaid with acoustic transducers to measure object curvature upon contact. Tanaka et al. \cite{tanaka2015tactile} devised an acoustic sensor for real-time lump detection in laparoscopic surgery, employing a silicone-based hollow tube and analysing sound wave deformations through the tube.

Chuang et al. \cite{chuang2018ultrasonic} designed an ultrasonic sensor capable of real-time static normal force measurements (1-6 N) and shape recognition, analyzing time-of-flight variations in ultrasonic pulses due to surface deformation. The sensor architecture includes a Thin Film Transistor layer sandwiched between piezoelectric PVDF layers and a soft polymer sensing surface. This design calls for sophisticated manufacturing. However, a revolution was brought by V. Wall et al. \cite{wall2022passive}, who extended acoustic methods to pinpoint the force location and material characteristics using a soft pneumatic finger equipped with an enclosed speaker and microphone. When featuring only a microphone, this design can still measure contact forces, locations, and materials, although it requires sound generation at the contact point for functionality \cite{zoller2018acoustic}.
These reports showcase the potential of using acoustic techniques to extract tactile information from soft material deformation. But for developing a low-cost tactile sensing skin, it is advisable (1) not to integrate complex electronic circuitry into the skin, and (2) relocating sensory hardware away from the sensing surface enables sensor compactness and avoids requiring sophisticated manufacturing techniques. Further, this allows for a variety of skin form factors. We made AST~\cite{rajendran2023towards, rajendran20232d} a low-cost and easily manufacturable standalone sensor. 


This paper presents a stand-alone and low-cost tactile sensing technology suitable for being used in different shapes and sizes, called Acoustic Soft Tactile skin (\textbf{ATS}) (Fig. \ref{fig:setup}a). 
Our contribution includes (i) a novel open-source low-cost tactile sensor that is easy to fabricate for fitting different shapes. A membrane with embedded acoustic channels, microphone, and speakers are the only key components of STS. (ii) We also present a design for a frame-less AST (\textbf{f-AST}) for non-flat surfaces. (iii) We investigated different regression and classification methods for learning the relation between the applied force and its location and the microphone readings for our AST designs during its performance testing.
 (iv) We demonstrate AST in robotic tasks where it is attached to a flat and non-flat surface and performs force-controlled robotic pick-and-place tasks. These tests prove that AST technology can be used in realistic robotic applications.  

\section{Acoustic Soft Tactile Skin}
\label{sec:AST}

\paragraph{Design:} To prove the proposed AST technology, we initially fabricated a flat, rectangular-shaped silicone skin measuring 35 mm x 60 mm. We mount the skin inside a 3D-printed casing to ensure portability and ease of testing and connecting the speaker-microphone unit. However, we have shown the skin design without a hard case in the later section of the paper. We investigated various skin configurations with single and dual ACs with simple geometrical shapes, such as cylindrical and conical, that run through the length of the skin. The diameter of the cylindrical AC is 5 mm, while the conical AC has diameters of 5 mm and 3 mm. The ACs connect the speaker-microphone arrangement of the skin. The prototyping procedure is detailed in the Appendix\footnote{\url{https://tinyurl.com/2mpy6vfr}}\label{myfootnote}.

The speaker unit generates continuous sound waves that travel through the ACs, and the microphone receives the sound waves. As the channels deform due to external physical interactions, the amplitude of the sound waves changes (see Fig.~\ref{fftvariation}), and we leverage ML models to capture the relationship between the amplitude changes and the tactile information. The details of the different skin configurations tested are presented in the following sections.

\begin{figure*}[tb!]   
\centering
    \begin{subfigure}[t]{0.15\textwidth}
    \centering
    \includegraphics[width=\textwidth]{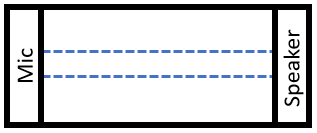}
    \caption{} \label{fig1}
\end{subfigure}
\begin{subfigure}[t]{0.15\textwidth}
    \centering
    \includegraphics[width=\textwidth]{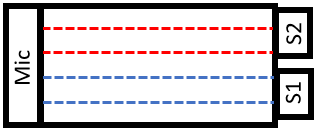}
    \caption{} \label{fig2}
\end{subfigure}
 \begin{subfigure}[t]{0.15\textwidth}
    \centering
    \includegraphics[width=\textwidth]{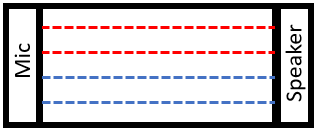}
    \caption{} \label{fig2b}
\end{subfigure}
 \begin{subfigure}[t]{0.15\textwidth}
    \centering
    \includegraphics[width=\textwidth]{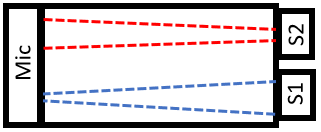}
    \caption{} \label{fig3a}
\end{subfigure}
 \begin{subfigure}[t]{0.15\textwidth}
    \centering
    \includegraphics[width=\textwidth]{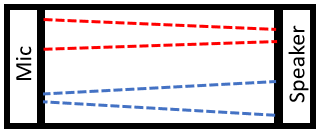}
    \caption{} \label{fig3b}
\end{subfigure}\\
 \begin{subfigure}[t]{0.15\textwidth}
    \centering
    \includegraphics[width=\textwidth]{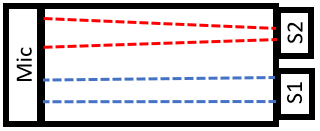}
    \caption{} \label{fig4a}
\end{subfigure}
 \begin{subfigure}[t]{0.15\textwidth}
    \centering
    \includegraphics[width=\textwidth]{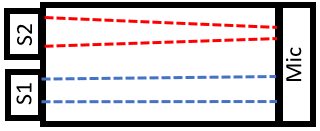}
    \caption{} \label{fig4b}
\end{subfigure}
\begin{subfigure}[t]{0.15\textwidth}
    \centering
    \includegraphics[width=\textwidth]{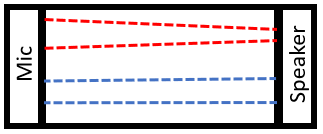}
    \caption{} \label{fig4c}
\end{subfigure}
\begin{subfigure}[t]{0.15\textwidth}
    \centering
    \includegraphics[width=\textwidth]{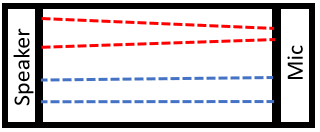}
    \caption{} \label{fig4d}
\end{subfigure}
\begin{subfigure}[t]{0.15\textwidth}
    \centering
    \includegraphics[width=\textwidth]{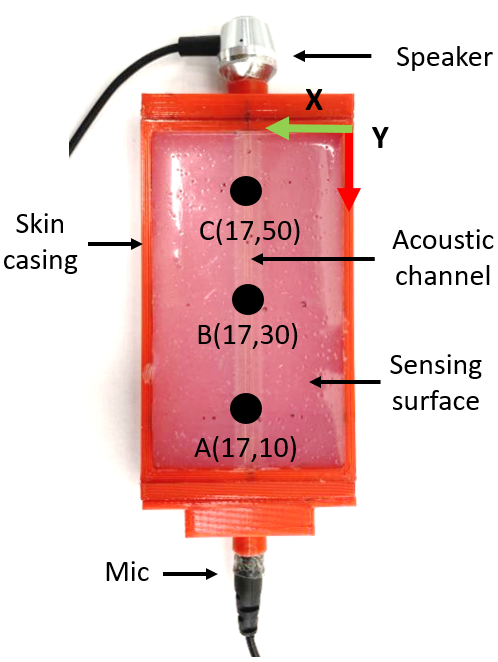}
    \caption{} \label{points}
\end{subfigure}
\caption{AST configurations: (a) AST 1 has a single channel; Double Channels: (b) AST 2a; (c) AST 2b; (d) AST 3a; (e) AST 3b; (f) AST 4a; (g) AST 4b; (h) AST 4c; (i) AST 4d; Calibration points selected for initial testing: $A = \{17,50\}$, $B= \{17,30\}$, and $C = \{17,10\}$ (S1 and S2 refer to two speakers)}
\vspace{-.3cm}
\label{sensorconfig}

\end{figure*}

\textbf{Acoustic Channel Design:} The impact of contact deformation on sound waves varies with the shape of the ACs, which we exploit to estimate force and contact location. We investigated the effect of different channel configurations on feature extraction for force and contact location (see Fig~\ref{sensorconfig} for AC designs).

We studied single-channel and dual-channel configurations to verify our hypothesis that ACs can be used to estimate tactile information. The study on single-channel skin configuration (AST 1) verifies the usability of this tactile skin, which calls for a narrow sensing region. For a broader sensing surface, the skin requires multiple channels, which we explore with studies on dual ACs (AST 2a-b, AST 3a-b, AST 4a-d) as an initial step. In future works, we will explore using multiple channels with different geometries spanning the entire skin area.
For single-channel skin (AST 1), we considered a cylindrical-shaped channel. For dual channel skin configurations, we have used combinations of (1) two cylindrical-shaped channels (AST 2a, AST 2b), (2) two conical-shaped channels (AST 3a, AST 3b), and (3) a conical-shaped channel with cylindrical-shaped one (AST 4a-4d). 

By providing these channel configurations, we also tested two other design hypotheses: (1) Can AC with a non-varying cross-section (cylinder) distinguish forces acting on different points along their length? (2) for dual-channel skins, can ACs with different geometrical shapes (AST 4a-4d) best capture the tactile information compared to skins with similar AC geometries (AST 2a-2b, AST 3a-3b)?


\textbf{Speaker Configurations:} Speaker configurations are the second primary design feature we investigated in this work. We used a single speaker for AST 1, AST 2b, AST 3b, AST 4c, and AST 4d. We provided an individual speaker for each channel for AST 2a, AST 3a, and AST 4a-4b.
\newline \noindent These speaker arrangements enable us to study the differences in skin performance (1) when all channels are provided with a single speaker versus each channel with an individual speaker and (2) when the speakers are arranged on the smaller and larger diameter ends of the conical channel (AST 4a and 4b, AST 4c and 4d).
In this study, we used computer headphone speakers and a microphone, which will be replaced with a miniature type in further development studies. 

\subsection{Extracting Tactile Feature from Sound Signal:}
We aim to estimate two tactile features, i.e., the normal force (Newtons) and their contact location from the sound wave's amplitude modulation. We use the data-driven approach to capture the mapping between the sound's wave amplitude modulation and these features. First, we created a dataset using a robot arm and load cell by applying force to a series of locations on the soft skin. We then used different regression and classification methods to learn the mapping and outputting of the tactile features. 
\section{Experiments and results}
\paragraph{Sensor Calibration:} We used a 6 DOF robot arm\footnote{The UFactory xArm by ufactory.cc} with a calibrated high-precision load cell (Miniature In Line Load Cell, a high-precision axial load cell with 0-1kN range and $10^{-5}$ N resolution) mounted on the robot wrist. It has an inbuilt driver board for USB communication with the PC and is mounted with a wedge-shaped 3D-printed peg as shown in Fig.~\ref{fig:setup}b.
We used the arm to apply a force on AST sensing points by the load cell and peg.  We initially considered three calibration points on the skin surface for all configurations, namely $A = \{17,50\}$, $B= \{17,30\}$, and $C = \{17,10\}$, as illustrated in Fig.~\ref{points}. Point A and C are closer to the microphone and the speaker, respectively. 

%

\begin{table*}[]
\centering
\small\addtolength{\tabcolsep}{-4pt}
\caption{Regression and Classification models selected for various skin configurations (SVM: Support Vector Machines, GP: Gaussian Process, NN: Neural Network, KNN: K-Nearest Neighbors)}
\label{selectedmodels}
\begin{tabular}{|l|l|c|l|c|}
\hline
\textbf{\begin{tabular}[c]{@{}l@{}}Skin\\ Configuration\end{tabular}} & \textbf{Regression model} & \textbf{RMSE} & \textbf{Classifier model} & \textbf{Accuracy (\%)} \\ \hline
AST 1                                                           & Exponential GP            & 0.72          & Bagged Ensemble Trees     & 96.4              \\
AST 2a                                                          & Rational Quadratic GP     & 2.21          & Fine Guassian SVM         & 95.7              \\
AST 2b                                                          & Matern 5/2 GP             & 3.24          & Weighted KNN              & 98.2              \\
AST 3a                                                          & Exponential GP            & 2.15          & Bilayered NN              & 92.5              \\
AST 3b                                                          & Exponential GP            & 1.06          & Bagged Ensemble Trees     & 91.9              \\
AST 4a                                                          & Exponential GP            & 3.25          & Weighted KNN              & 92.8              \\
AST 4b                                                          & Exponential GP            & 3.6           & Fine Gaussian SVM         & 95.4              \\
AST 4c                                                          & Exponential GP            & 2.53          & Weighted KNN              & 97.2              \\
AST 4d                                                          & Exponential GP            & 1.18          & Weighted KNN              & 97.3              \\ \hline
\end{tabular}
\end{table*}

For the calibration process, the AST was positioned in a fixed location on the workbench, as depicted in Fig. \ref{fig:setup}b, and a continuous reference sound signal was emitted through the speaker(s). The specific characteristics of this reference signal are outlined in the upcoming subsection. Subsequently, the robot arm sequentially approached each calibration point while a peg attached to the arm exerted pressure by an incremental push move of 0.2 mm. As the peg's central axis traversed vertically through the designated points, the AC(s) underwent compression. At every 0.2 mm increment, the corresponding load cell measurements were recorded, and simultaneously, the microphone captured 50 samples of the received sound signal. This procedure was repeated until the load cell reading reached 30$^{+1}$ N ($^*$, $*$ indicates the tolerance of the value).


\textbf{Reference Sound Signal:}
\label{refsignal}
A test reference sound signal is generated using Audacity. It comprises four sine waves of frequencies of 300, 500, 700, and 900 Hz with an amplitude of 0.6 (on a 0 to 1 scale). During the sensor operation, this reference signal is played through the speaker(s).

\textbf{Data Processing:}
The data processing pipeline for the calibration process is shown in Fig.~\ref{fig:setup}b. After data processing, the resulting dataset contains the load cell reading with the corresponding contact location (A, B, or C) and the modulated sound waves' amplitudes and frequencies. We use Fast Fourier Transform (\textbf{FFT}) to compute the amplitudes of the sound signals received by the microphone (FFT data). For each skin model, 5100 data points are generated, with 1700 data points each for the three locations, A, B, and C.

\subsection{Force and Contact Location Estimation:}
\begin{figure*}[tb!]   
\centering
\begin{subfigure}[t]{0.245\textwidth}
\includegraphics[width=\textwidth]{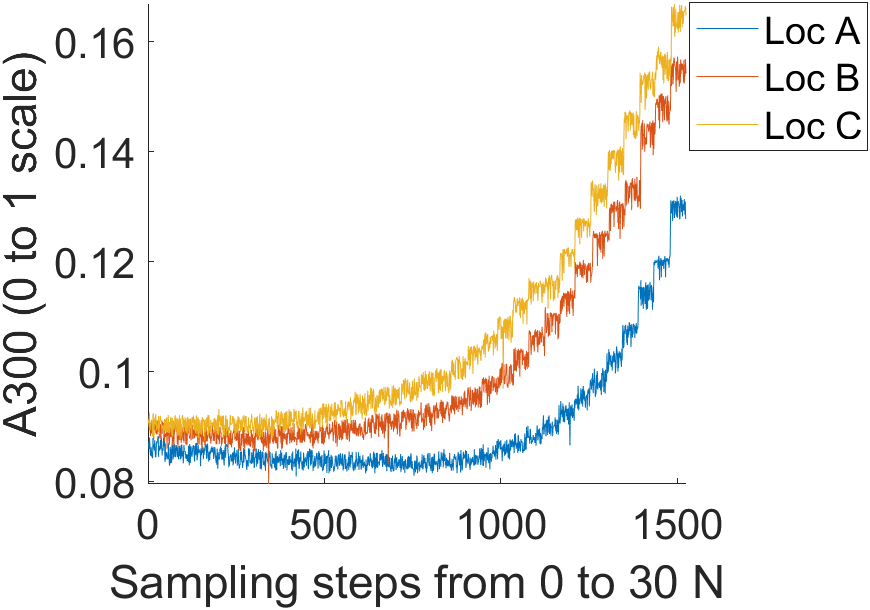}
\caption{300 Hz component} \label{A300}
\end{subfigure}
\begin{subfigure}[t]{0.24\textwidth}   
    \includegraphics[width=\textwidth]{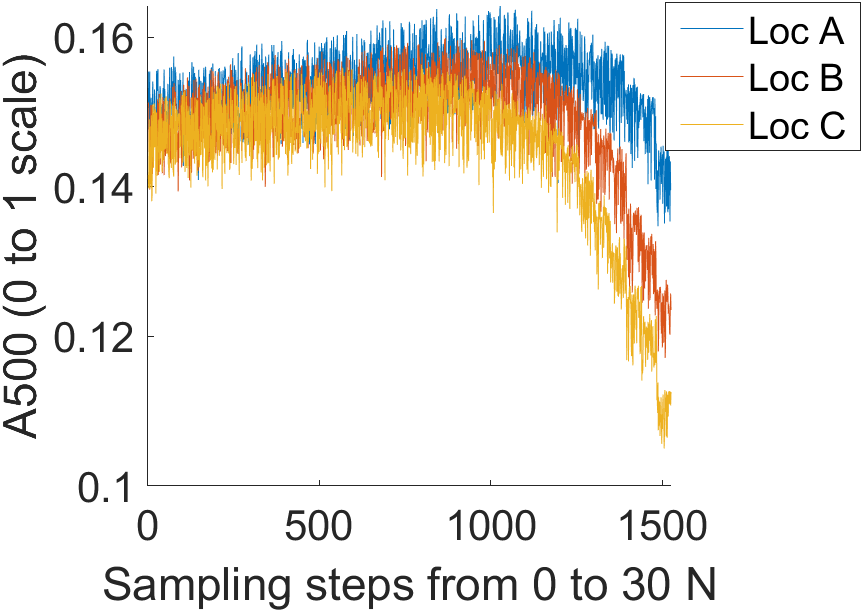}
    \caption{500 Hz component} \label{A500}
\end{subfigure}
\begin{subfigure}[t]{0.24\textwidth}  
\includegraphics[width=\textwidth]{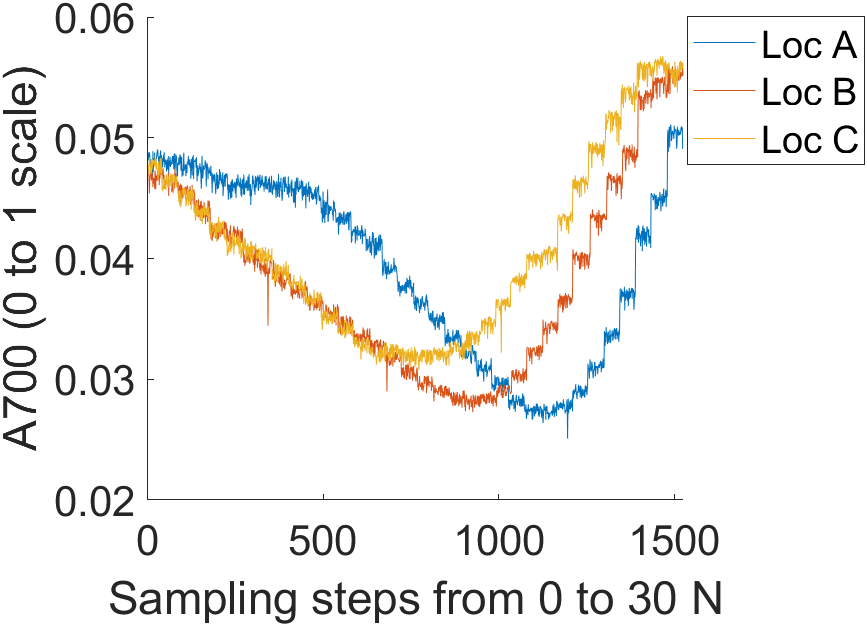}
\caption{700 Hz component} \label{A700}
\end{subfigure}
\begin{subfigure}[t]{0.24\textwidth}
    \includegraphics[width=\textwidth]{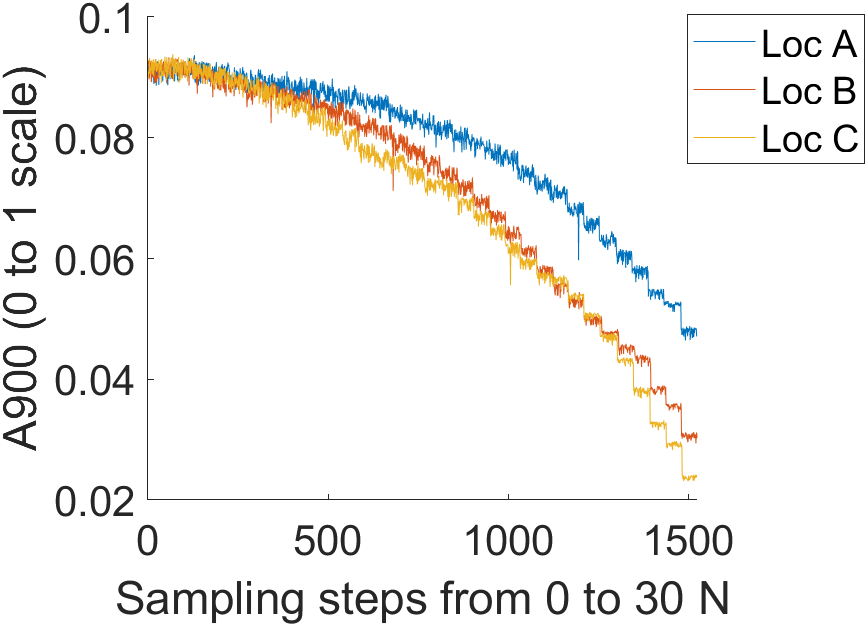}
    \caption{900 Hz component} \label{A900}
\end{subfigure}
\caption{Variation of FFT data at locations A, B, and C when force varies from 0 to 30$^{+1}$ N}
\label{fftvariation}
\vspace{-.4cm}
\end{figure*}

We learned the amplitudes of individual frequency components in the reference signal (FFT data) vary with an increase in force from 0 to 30$^{+1}$ N at each location A, B, and C on the STS. Fig.~\ref{fftvariation} presents a sample of these variations for AST 1. This indicates the collected FFT data effectively estimate unknown forces and their contact locations. Therefore, we express the unknown force and its location as:
\begin{equation}
(F_i, L_i) = f(A_{3, i}, A_{5, i}, A_{7, i}, A_{9, i})
\end{equation} where $F_i$ is the unknown force at instance $i$, $A_{3, i}, A_{5, i}, A_{7, i}, A_{9, i}$ are the amplitudes of frequency components of the FFT at 300, 500, 700, and 900 Hz recorded at the same instance (FFT data), and $L_i$ is the location of contact (A, B or C).

\textbf{Performance Metrics:}
The following metrics present the outcomes of the tests conducted to present the findings. 
(i) We use validation error and/or efficiency for the selection of regression and classifier models and each STS.
 (ii) The performance force estimation is presented as the percentage of estimations that fall within $\pm$0.50 N, $\pm$1 N, and $\pm$1.50 N tolerance of the actual force values.
(iii) The contact location estimation performance is presented as the total number of true estimations per test (170 trials are tested for each AST location), and they are averaged to define the overall accuracy.
We compared various regression and classifier models (see Appendix\footnotemark[1]) to estimate the force and its location from the FFT data. We compare these models' validation errors/accuracy for each AST data. The models are trained using a dataset partition of 90:10 and 10-fold cross-validation. The \emph{regression} model with the \emph{minimum validation error} and the \emph{classifier} with the \emph{maximum accuracy} are selected for estimating force and location, respectively (refer to table~\ref {selectedmodels}). The details of force and location estimations using these models are presented below.

\begin{figure*}[tb!]   
\centering
    \begin{subfigure}[t]{0.25\textwidth}
    \includegraphics[width=\textwidth]{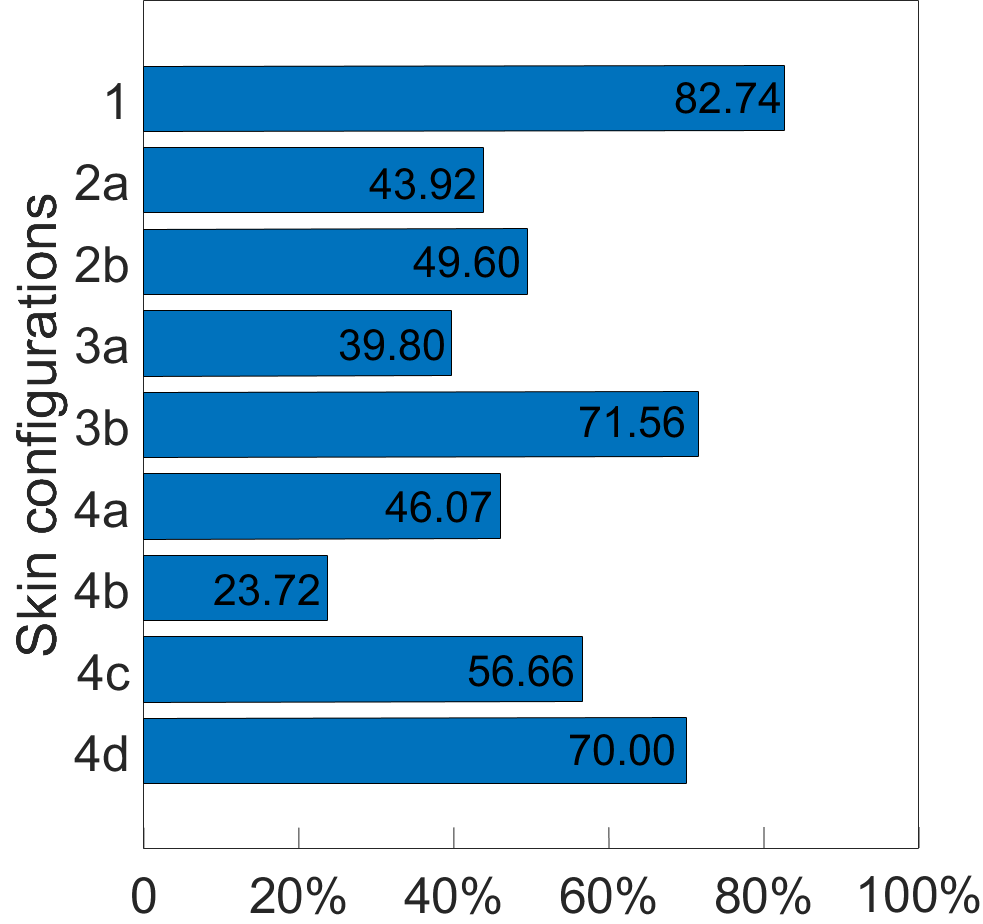}
    \caption{$\pm$ 0.50 N tolerance} \label{fig9a}
\end{subfigure} \hspace{1cm}
\begin{subfigure}[t]{0.25\textwidth}
    
    \includegraphics[width=\textwidth]{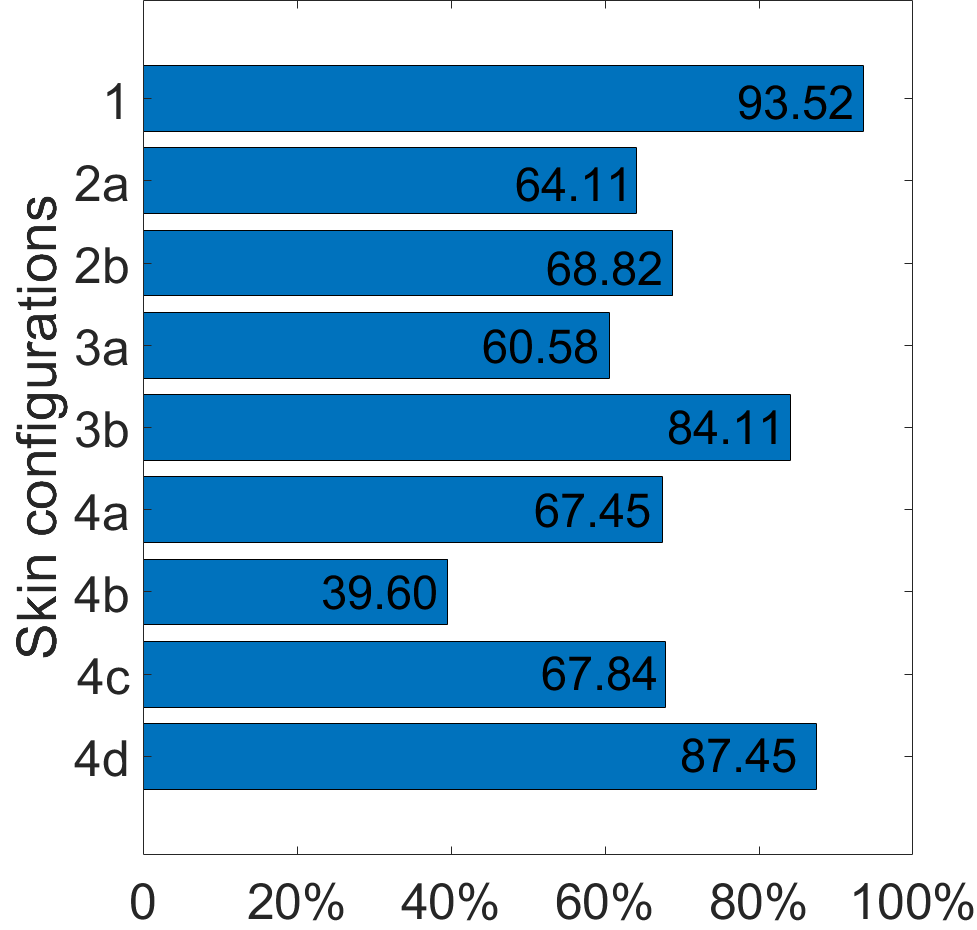}
    \caption{$\pm$ 1 N tolerance} \label{fig9b}
\end{subfigure} \hspace{1cm}
 \begin{subfigure}[t]{0.25\textwidth}
    
    \includegraphics[width=\textwidth]{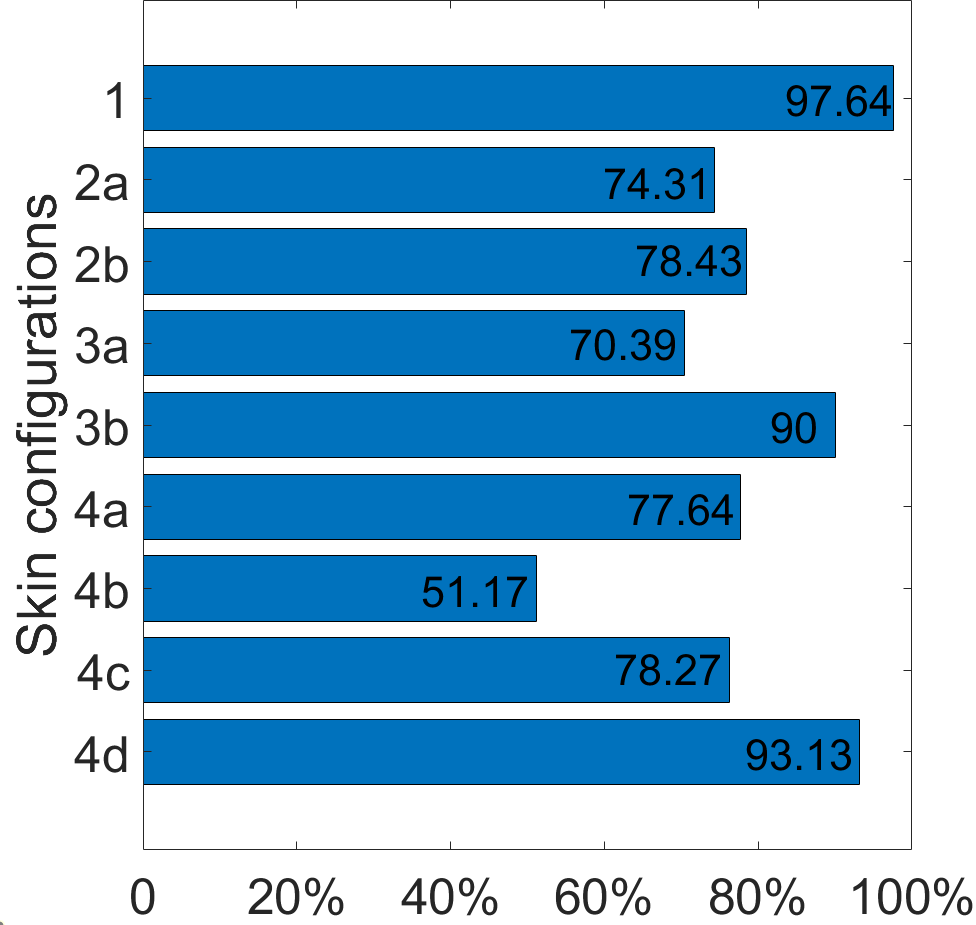}
    \caption{$\pm$ 1.50 N tolerance} \label{fig9c}
\end{subfigure}
\caption{Performance of different configurations of STS: percentage of force estimated with $\pm$ 0.50 N (a), $\pm$ 1 N (b), and $\pm$ 1.50 N (c) tolerance.}
\label{ForceResult1}
\vspace{-.4cm}
\end{figure*}

\textbf{Force estimation:}
The force estimation performance of the STSs is presented in Fig.~\ref{ForceResult1}, showing the percentage of force estimations falling within $\pm$0.5 N, $\pm$1 N, and $\pm$1.5 N tolerances. The performance of each skin configuration is analysed based on the effect of AC geometries and speaker configuration.
\begin{table*}[t!]
\centering
\small\addtolength{\tabcolsep}{-4pt}
\caption{ Estimating the location of contact force for 170 test cases for  points A, B, C as shown in Fig.~\ref{points}}
\label{contactprediction}
\begin{tabular}{|c|ccccccccc|}

\hline
\multirow{2}{*}{\textbf{True predictions}} & \multicolumn{9}{c|}{\textbf{STS configuration}}                                                                                                                                                                                                                                       \\ \cline{2-10} 
                     & \multicolumn{1}{c|}{STS 1}    & \multicolumn{1}{c|}{STS 2a}    & \multicolumn{1}{c|}{STS 2b}    & \multicolumn{1}{c|}{STS 3a}    & \multicolumn{1}{c|}{STS 3b}    & \multicolumn{1}{c|}{STS 4a}    & \multicolumn{1}{c|}{STS 4b}    & \multicolumn{1}{c|}{STS 4c}    & \multicolumn{1}{c|}{STS 4d} \\ \hline
A                    & \multicolumn{1}{c|}{169}   & \multicolumn{1}{c|}{169}   & \multicolumn{1}{c|}{168}   & \multicolumn{1}{c|}{165}   & \multicolumn{1}{c|}{163}   & \multicolumn{1}{c|}{168}   & \multicolumn{1}{c|}{168}   & \multicolumn{1}{c|}{168}   & 168   \\ \hline
B                   & \multicolumn{1}{c|}{160}   & \multicolumn{1}{c|}{168}   & \multicolumn{1}{c|}{169}   & \multicolumn{1}{c|}{150}   & \multicolumn{1}{c|}{142}   & \multicolumn{1}{c|}{151}   & \multicolumn{1}{c|}{156}   & \multicolumn{1}{c|}{165}   & 167   \\ \hline
C                   & \multicolumn{1}{c|}{164}   & \multicolumn{1}{c|}{153}   & \multicolumn{1}{c|}{168}   & \multicolumn{1}{c|}{161}   & \multicolumn{1}{c|}{150}   & \multicolumn{1}{c|}{155}   & \multicolumn{1}{c|}{162}   & \multicolumn{1}{c|}{166}   & 162   \\ \hline    
\textbf{Accuracy} (\%)                    & \multicolumn{1}{c|}{96.66} & \multicolumn{1}{c|}{96.07} & \multicolumn{1}{c|}{99.01} & \multicolumn{1}{c|}{93.33} & \multicolumn{1}{c|}{89.21} & \multicolumn{1}{c|}{92.94} & \multicolumn{1}{c|}{95.29} & \multicolumn{1}{c|}{97.84} & 97.45 \\ \hline
\end{tabular}
\end{table*}

\subsubsection{Effect of Acoustic Channel Geometry}
The single-channel AST 1 configuration demonstrated a force estimation accuracy of 82.74\% within $\pm$0.5 N tolerance range and 93.5\% within $\pm$1 N range, indicating that a single channel with a uniform geometrical shape can accurately infer contact forces acting on different points. Other skin configurations were also tested to investigate the feasibility of using an array of AC to serve a skin that may require a broader sensing surface area. The results in Fig.~\ref{ForceResult1} show that considerably better performance can be achieved when the channel geometries are non-identical and a single speaker unit serves all channels. This is evident when comparing the performance of skin configuration AST 3b, AST 4c, and AST 4d with AST 2b.

\subsubsection{Effect of Speaker Configuration}
The results in Fig~\ref{ForceResult1} indicate that the dual-channel skin configurations with a single speaker (AST 2b, AST 3b, AST 4c, AST 4d) mostly outperformed others or showed comparable performance to their speaker designs (AST 2a, AST 3a, AST 4a, AST 4b). This suggests that a single speaker can serve multiple channels without compromising performance. This finding can imply that the skin technology is potentially scalable, and a skin with multiple channels may not require an equal number of speakers (this is an open question for future works). 
Speakers on different ends of channels did not show any significant impact on the sensor performance.


\textbf{Contact Location Estimation:}
Table~\ref{contactprediction} presents the accuracy of each AST configuration for estimating the contact locations (tested for 170 cases per location). All skin configurations achieved a high accuracy of $<$89\% for location estimation. The results in Table~\ref{contactprediction} also indicate that the performance of location estimation is better with a single speaker serving the AC(s).  

The results demonstrate (1) A smaller-width skin can utilise a single acoustic channel to measure contact force and its location accurately. The channel can also have a simple geometry, such as a cylinder. Although the shape is fixed along its length, it can still distinguish forces and their locations applied at different points. (2) For skin with a broader sensing surface area, an array of non-identical ACs can be used, and individual speakers for each channel may not be necessary. (3) Further study will investigate AST with different AC shapes that may lead to improved AST performance.


\section{Frame-less STS}
\begin{figure*}[tb!]   
\centering

\begin{subfigure}[t]{0.2\textwidth}
    \includegraphics[width=\textwidth]{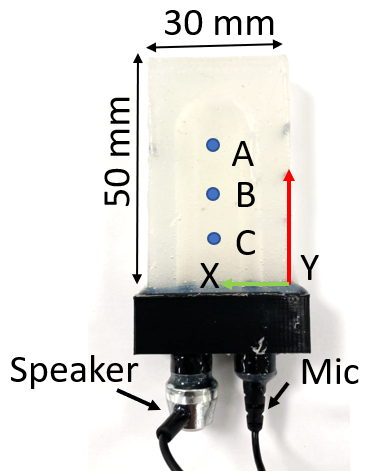}
    \caption{} 
    \label{asfl4}
\end{subfigure} \hspace{.41cm}
 \begin{subfigure}[t]{0.20\textwidth}
    \includegraphics[width=\textwidth]{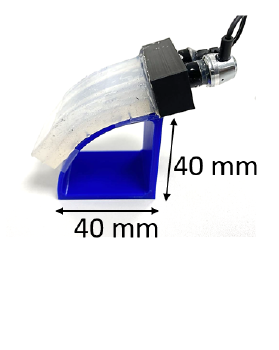}
    \caption{} \label{asfl5}
\end{subfigure} \hspace{.41cm}
    \begin{subfigure}[t]{0.25\textwidth}  
    \includegraphics[width=\textwidth]{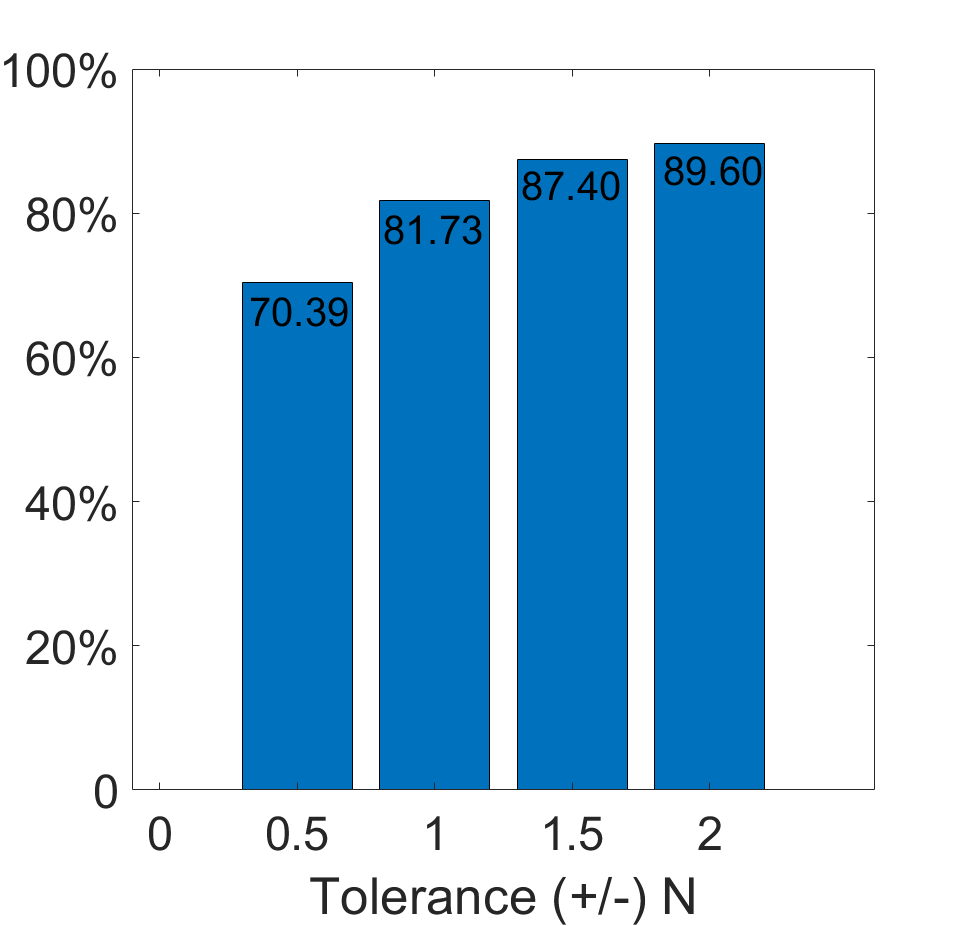}
    \caption{} \label{astflforce}
\end{subfigure} \hspace{.41cm}
\begin{subfigure}[t]{0.18\textwidth} 
    \includegraphics[width=\textwidth]{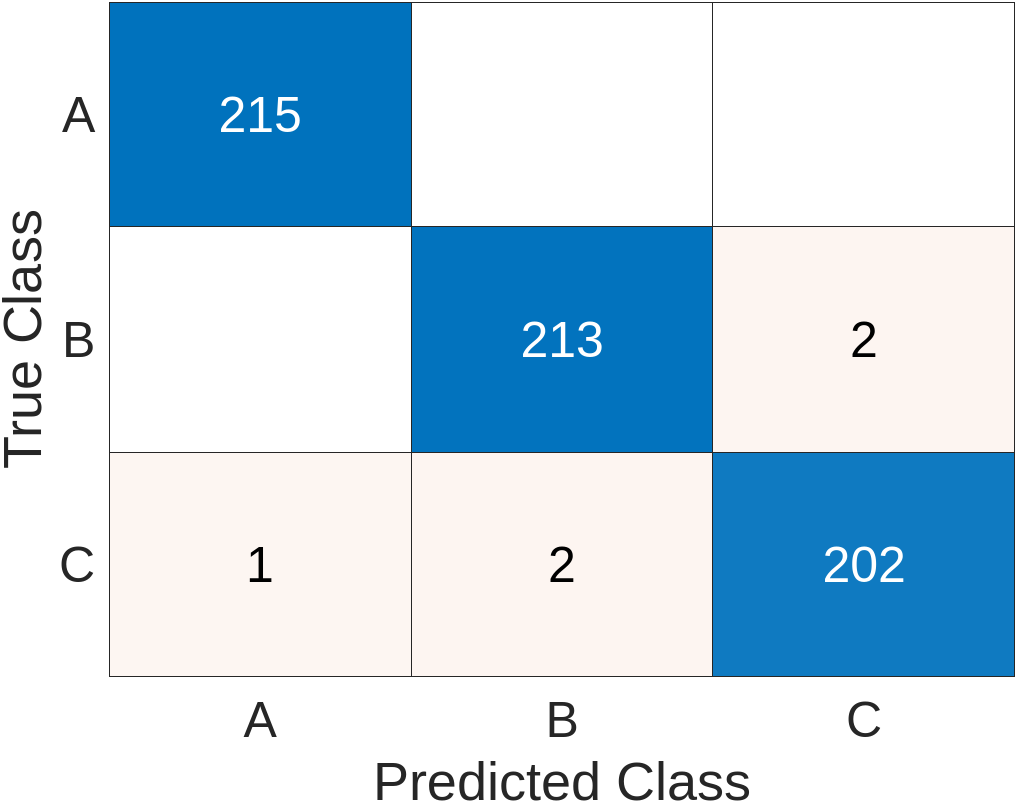}
    \caption{} \label{astflloc}
\end{subfigure}
\caption{Frame-less AST: (a) Frame-less AST (f-AST) with calibration points: $A = \{15,35\}$, $B= \{15,25\}$, and $C = \{15,15\}$ (b) skin mounted on the curved surface of a finger; (c); (d) Percentage of force estimations with $\pm$ 0.50 N, $\pm$ 1 N, and $\pm$ 1.50 N tolerance; (d) Contact location true predictions. }
\label{astfldesign}
\vspace{-.4cm}
\end{figure*}

Frame-less AST (f-AST) is the evolution of this AST concept, where it has a flexible sensing membrane capable of being fastened to any curved surface. This skin design can provide a broader range of robotic equipment with tactile sensation, such as enabling safer human-robot interactions in shared work environments. The test design of f-AST has a casing to house the speaker and microphone at the bottom and from which the sensing surface extends (refer fig.\ref{asfl4}). The acoustic channel design is cylindrical, connecting the speaker and microphone, inspired by AST 1 design which was the best-performing configuration. To test its capability of being used on curved surfaces for tactile sensing, the f-AST is attached to a 3D-printed curved surface, as shown in fig.\ref{asfl5}. The prototyping procedure is detailed in the Appendix\footnotemark[1]. 

To test the force and contact location capability, we employed a similar calibration process as used before. We chose three calibration points spaced 10 mm apart (while higher resolution calibration is feasible, it requires significantly more time) as illustrated in Fig.\ref{asfl4}. A total of 6350 data points were collected, and we utilised a \emph{90:10} data partitioning scheme for training and testing purposes. Here too, we evaluated various regression and classification methods, ultimately selecting the optimal model for estimating the force and its location. The Bagged Trees Ensemble exhibited the most promising cross-validation results, yielding a cross-validation error of 1.77 N for force estimation and a cross-validation accuracy of 99.2\% for estimating force location.
%
Fig.\ref{astflforce} presents the percentage of accurately estimated forces with different tolerance values. Notably, 70.39\% of force estimations fall within the $\pm$ 0.50 N tolerance range, while this percentage increases to 81.73\%, 87.40\%, and 89.60\% within $\pm$ 1 N, $\pm$ 1.50 N and $\pm$ 2 N tolerances, respectively. AST demonstrates an impressive 99\% accuracy for contact location estimation (Fig.~\ref{astflloc}). These results show AST technology can be used as a sensory cover for surfaces with different geometry needing tactile sensing. 


%
\begin{table}[]
    \centering
    \small\addtolength{\tabcolsep}{-4pt}
    \caption{Grip force control with frame-less STS. The values in the table are the Mean Absolute Error (MAE) of the measured grip force by AST with respect to the desired grip force. The lower the MAE, the better the performance of the controller.} 
    \begin{tabular}{| c | c | c | c | c | c | c |}
    \hline
          \multirow{2}{*}{\begin{tabular}{@{}c@{}}Target \\ Grip Force\end{tabular}}  & \multirow{2}{*}{\begin{tabular}{@{}c@{}}White \\ Noise\end{tabular}} & \multicolumn{5}{|c|}{Mean Absolute Error} \\
          \cline{3-7}
          & & Chess & Dice & Lemon & Marble & Strawberry  \\ 
          \hline 
         \multirow{2}{*}{2 N} & Off & 0.03 & 0.04 & 0.04 & \textbf{0.05} & 0.02  \\
          & On & 0.08 & 0.05 & 0.05 & 0.06 & \textbf{0.09}  \\ \hline
         \multirow{2}{*}{10 N} & Off & 0.26 & 0.14 & 0.38 & 0.40 & \textbf{0.51}  \\
          & On & 0.36 & 0.22 & 0.38 & 0.78 & \textbf{0.91}  \\  \hline
          \multirow{2}{*}{20 N} & Off & 0.40 & \textbf{1.20} & 0.40 & 0.73 & 0.19  \\
          & On & 0.61 & \textbf{1.10} & 0.63 & 0.81 & 0.21  \\
         \hline
    \end{tabular}
    \label{tab:grip_ctl}
    \vspace{-.3cm}
\end{table}
\begin{figure}[tb!]
\centering
\begin{subfigure}{0.235\textwidth}
    \includegraphics[width=\linewidth]{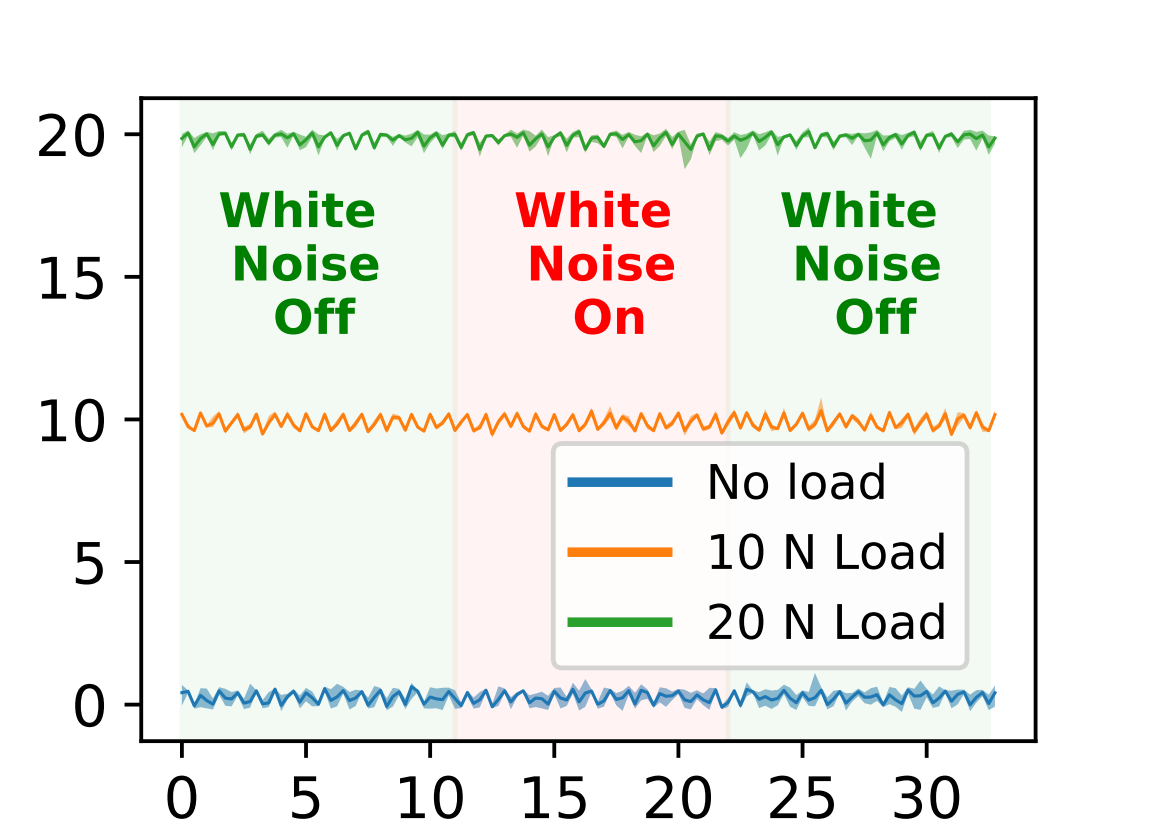}
    \caption{} 
    \label{fig:whitenoise}
\end{subfigure}
 \begin{subfigure}{0.23\textwidth}
    \includegraphics[width=\linewidth]{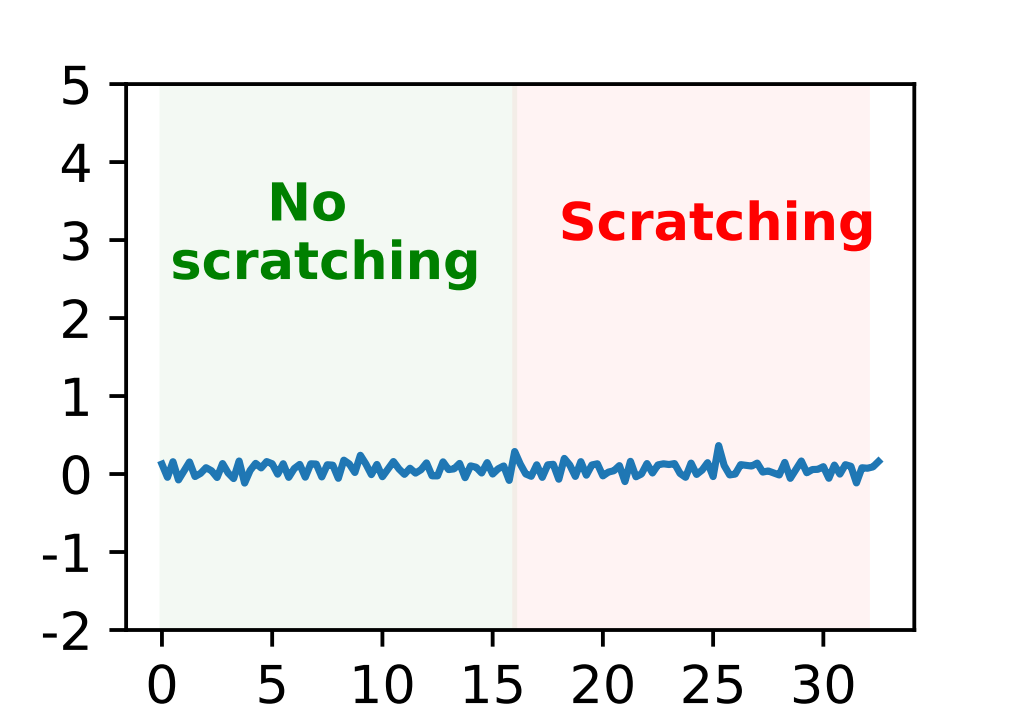}
    \caption{} \label{fig:scratch}
\end{subfigure}
\caption{Noise interference tests (Force [N] vs time [s]). (a) AST readings in the presence of white noise under no load and 10 N and 20 N loading conditions. (b) AST readings when performing scratching on the sensor surface with a stiff paintbrush.}
\label{fig:noise}
\vspace{-0.6cm}
\end{figure}
\subsection{Noise Interference Test}
We study the robustness of f-AST in the presence of white noise at 100 decibels for zero, 10 N, and 20 N constant loading on point B.
Fig~\ref{fig:whitenoise} shows the sensor readings in the white noise tests.
In test trials, we start with 11 seconds of normal conditions, then turn on the noise for the next 11 seconds, and finally return to normal conditions for the last 11 seconds. We repeated each loading test 5 times. Fig.~\ref{fig:whitenoise} shows the mean and variance of the f-AST reading, demonstrating it is robust to the white ambient noise. We also tested the effect of the contact noise on f-AST readings by scratching the f-AST surface with no load condition with a stiff paintbrush for 11 seconds (Fig.\ref{fig:scratch}). The figure shows that the sensor readings remain constant for smooth strokes compared to the normal no-load condition. Testing the contact noise effect under different loading conditions can be complex, as distinguishing the changes in f-AST readings resulting from contact noise or from the actual contact forces may not be trivial.

\subsection{Robotic Application}
We tested the f-AST in a pick-and-place task with real-time grip force control. We evaluated the performance of the controller by gripping five test objects from the YCB object set~\cite{calli2015benchmarking} with and without the presence of white noise (100 decibels). Fig.~\ref{fig:setup}c shows an experimental setup consisting of (1) a seven dof Franka Emika arm, (2) an SMC LEZH gripper, and (3) an f-AST. SMC gripper has a 30 mm stroke with a maximum 210 N grip force. The control task includes (i) gripping an object from a top pose with a specified grip force, (ii) lifting the object upward 10 cm, (iii) moving it to a goal point, and (iv) dropping it there. The grip force controller aims to keep the grip force constant during stages using the AST tactile feedback. When the robot reaches the grip pose, the gripper starts decreasing its grip width with a 1mm step size until the measured grip force by f-AST sensor reaches a target desired value. Then, the lift, move, and drop stages are executed automatically. We tested the controller for three target grip forces, namely 2N, 10N, and 20N.

Table~\ref{tab:grip_ctl} shows the Mean Absolute Error (MAE) of the target grip forces per object, and they are in the acceptable range for an early-stage prototype. The target force is based on the f-AST sensor readings and not a true contact force from a load cell. For 2N and 10N target grip forces, strawberry resulted in the highest MAE. For 20N, the dice show the highest MAE. This is because these contact shapes are not in the calibration data set. However, this can be covered by an extensive calibration dataset. The table also shows that although the MAE in the reading during the white noise can be slightly higher than the normal test condition, they are very close to the desired force value.

\section{Conclusion}
\label{sec:conclusion}

We presented the concept of a novel Soft Tactile skin technology based on sound wave amplitude modulation. AST uses Acoustic Channels beneath the sensing surface to measure the contact normal force and its location. To validate the concept, we studied different AST designs with different configurations of Acoustic Channels, acoustic hardware (speaker-mic) to validate the concept. During our study, AST made more than 93\% of static normal force estimations with $\pm$1.5 N tolerance for a full-scale force range of 0-30$^{+1}$ N and also made more than 96\% accurate force contact location estimations.
Moreover, we made a prototype of a frame-less AST (f-AST) and tested it. f-AST shows the potential of using the AST technology as a tactile sensory cover on curved surfaces and other real-time robotic applications. Overall, the AST has the potential to become a valuable tool for tactile measurements for various robotic applications. We hope AST contributes to the advances in learning physical robot interactions.




\bibliographystyle{IEEEtran}

\bibliography{IEEEexample,Reference}

\begin{thebibliography}{10}
\providecommand{\url}[1]{#1}
\csname url@samestyle\endcsname
\providecommand{\newblock}{\relax}
\providecommand{\bibinfo}[2]{#2}
\providecommand{\BIBentrySTDinterwordspacing}{\spaceskip=0pt\relax}
\providecommand{\BIBentryALTinterwordstretchfactor}{4}
\providecommand{\BIBentryALTinterwordspacing}{\spaceskip=\fontdimen2\font plus
\BIBentryALTinterwordstretchfactor\fontdimen3\font minus
  \fontdimen4\font\relax}
\providecommand{\BIBforeignlanguage}[2]{{%
\expandafter\ifx\csname l@#1\endcsname\relax
\typeout{** WARNING: IEEEtran.bst: No hyphenation pattern has been}%
\typeout{** loaded for the language `#1'. Using the pattern for}%
\typeout{** the default language instead.}%
\else
\language=\csname l@#1\endcsname
\fi
#2}}
\providecommand{\BIBdecl}{\relax}
\BIBdecl

\bibitem{mandil2023tactile}
W.~Mandil, V.~Rajendran, K.~Nazari, and A.~Ghalamzan-Esfahani,
  ``Tactile-sensing technologies: Trends, challenges and outlook in agri-food
  manipulation,'' \emph{Sensors}, vol.~23, no.~17, p. 7362, 2023.

\bibitem{yuan2017gelsight}
W.~Yuan, S.~Dong, and E.~H. Adelson, ``Gelsight: High-resolution robot tactile
  sensors for estimating geometry and force,'' \emph{Sensors}, vol.~17, no.~12,
  p. 2762, 2017.

\bibitem{lepora2021soft}
N.~F. Lepora, ``Soft biomimetic optical tactile sensing with the tactip: A
  review,'' \emph{IEEE Sensors Journal}, vol.~21, no.~19, pp. 21\,131--21\,143,
  2021.

\bibitem{fritzsche2011tactile}
M.~Fritzsche, N.~Elkmann, and E.~Schulenburg, ``Tactile sensing: A key
  technology for safe physical human robot interaction,'' in \emph{Proceedings
  of the 6th International Conference on Human-robot Interaction}, 2011, pp.
  139--140.

\bibitem{yousef2011tactile}
H.~Yousef, M.~Boukallel, and K.~Althoefer, ``Tactile sensing for dexterous
  in-hand manipulation in robotics—a review,'' \emph{Sensors and Actuators A:
  physical}, vol. 167, no.~2, pp. 171--187, 2011.

\bibitem{nazari2021tactile}
K.~Nazari, W.~Mandill, M.~Hanheide, and A.~G. Esfahani, ``Tactile dynamic
  behaviour prediction based on robot action,'' in \emph{Towards Autonomous
  Robotic Systems: 22nd Annual Conference, TAROS 2021, Lincoln, UK, September
  8--10, 2021, Proceedings 22}.\hskip 1em plus 0.5em minus 0.4em\relax
  Springer, 2021, pp. 284--293.

\bibitem{Mandil2022RSS}
W.~Mandil, K.~Nazari, and A.~Ghalamzan, ``{Action Conditioned Tactile
  Prediction: case study on slip prediction},'' in \emph{Proceedings of
  Robotics: Science and Systems}, New York City, NY, USA, 6 2022.

\bibitem{nazari2023proactive}
K.~Nazari, W.~Mandil, and A.~M.~G. Esfahani, ``Proactive slip control by
  learned slip model and trajectory adaptation,'' in \emph{Conference on Robot
  Learning}.\hskip 1em plus 0.5em minus 0.4em\relax PMLR, 2023, pp. 751--761.

\bibitem{nazari2023deep}
K.~Nazari, G.~Gandolfi, Z.~Talebpour, V.~Rajendran, P.~Rocco \emph{et~al.},
  ``Deep functional predictive control for strawberry cluster manipulation
  using tactile prediction,'' \emph{arXiv preprint arXiv:2303.05393}, 2023.

\bibitem{dahiya2009tactile}
R.~S. Dahiya, G.~Metta, M.~Valle, and G.~Sandini, ``Tactile sensing—from
  humans to humanoids,'' \emph{IEEE transactions on robotics}, vol.~26, no.~1,
  pp. 1--20, 2009.

\bibitem{dargahi2005advances}
J.~Dargahi and S.~Najarian, ``Advances in tactile sensors design/manufacturing
  and its impact on robotics applications--a review,'' \emph{Industrial Robot:
  An International Journal}, 2005.

\bibitem{zimmer2019predicting}
J.~Zimmer, T.~Hellebrekers, T.~Asfour, C.~Majidi, and O.~Kroemer, ``Predicting
  grasp success with a soft sensing skin and shape-memory actuated gripper,''
  in \emph{2019 IEEE/RSJ International Conference on Intelligent Robots and
  Systems (IROS)}.\hskip 1em plus 0.5em minus 0.4em\relax IEEE, 2019, pp.
  7120--7127.

\bibitem{li2016wide}
Q.~Li, Z.~Ullah, W.~Li, Y.~Guo, J.~Xu, R.~Wang, Q.~Zeng, M.~Chen, C.~Liu, and
  L.~Liu, ``Wide-range strain sensors based on highly transparent and supremely
  stretchable graphene/ag-nanowires hybrid structures,'' \emph{Small}, vol.~12,
  no.~36, pp. 5058--5065, 2016.

\bibitem{song2019pneumatic}
K.~Song, S.~H. Kim, S.~Jin, S.~Kim, S.~Lee, J.-S. Kim, J.-M. Park, and Y.~Cha,
  ``Pneumatic actuator and flexible piezoelectric sensor for soft virtual
  reality glove system,'' \emph{Scientific reports}, vol.~9, no.~1, p. 8988,
  2019.

\bibitem{rehan2022soft}
M.~Rehan, M.~M. Saleem, M.~I. Tiwana, R.~I. Shakoor, and R.~Cheung, ``A soft
  multi-axis high force range magnetic tactile sensor for force feedback in
  robotic surgical systems,'' \emph{Sensors}, vol.~22, no.~9, p. 3500, 2022.

\bibitem{diguet2022tactile}
G.~Diguet, J.~Froemel, M.~Muroyama, and K.~Ohtaka, ``Tactile sensing using
  magnetic foam,'' \emph{Polymers}, vol.~14, no.~4, p. 834, 2022.

\bibitem{wu2022new}
H.~Wu, B.~Zheng, H.~Wang, and J.~Ye, ``New flexible tactile sensor based on
  electrical impedance tomography,'' \emph{Micromachines}, vol.~13, no.~2, p.
  185, 2022.

\bibitem{ward2018tactip}
B.~Ward-Cherrier, N.~Pestell, L.~Cramphorn, B.~Winstone, M.~E. Giannaccini,
  J.~Rossiter, and N.~F. Lepora, ``The tactip family: Soft optical tactile
  sensors with 3d-printed biomimetic morphologies,'' \emph{Soft robotics},
  vol.~5, no.~2, pp. 216--227, 2018.

\bibitem{gomes2022geltip}
D.~F. Gomes and S.~Luo, ``Geltip tactile sensor for dexterous manipulation in
  clutter,'' in \emph{Tactile Sensing, Skill Learning, and Robotic Dexterous
  Manipulation}.\hskip 1em plus 0.5em minus 0.4em\relax Elsevier, 2022, pp.
  3--21.

\bibitem{donlon2018gelslim}
E.~Donlon, S.~Dong, M.~Liu, J.~Li, E.~Adelson, and A.~Rodriguez, ``Gelslim: A
  high-resolution, compact, robust, and calibrated tactile-sensing finger,'' in
  \emph{2018 IEEE/RSJ International Conference on Intelligent Robots and
  Systems (IROS)}.\hskip 1em plus 0.5em minus 0.4em\relax IEEE, 2018, pp.
  1927--1934.

\bibitem{lambeta2020digit}
M.~Lambeta, P.-W. Chou, S.~Tian, B.~Yang, B.~Maloon, V.~R. Most, D.~Stroud,
  R.~Santos, A.~Byagowi, G.~Kammerer \emph{et~al.}, ``Digit: A novel design for
  a low-cost compact high-resolution tactile sensor with application to in-hand
  manipulation,'' \emph{IEEE Robotics and Automation Letters}, vol.~5, no.~3,
  pp. 3838--3845, 2020.

\bibitem{10017344}
Z.~Chen, S.~Zhang, S.~Luo, F.~Sun, and B.~Fang, ``Tacchi: A pluggable and low
  computational cost elastomer deformation simulator for optical tactile
  sensors,'' \emph{IEEE Robotics and Automation Letters}, vol.~8, no.~3, pp.
  1239--1246, 2023.

\bibitem{sferrazza2019design}
C.~Sferrazza and R.~D’Andrea, ``Design, motivation and evaluation of a
  full-resolution optical tactile sensor,'' \emph{Sensors}, vol.~19, no.~4, p.
  928, 2019.

\bibitem{gong2017pneumatic}
D.~Gong, R.~He, J.~Yu, and G.~Zuo, ``A pneumatic tactile sensor for
  co-operative robots,'' \emph{Sensors}, vol.~17, no.~11, p. 2592, 2017.

\bibitem{chuang2018ultrasonic}
C.-H. Chuang, H.-K. Weng, J.-W. Chen, and M.~O. Shaikh, ``Ultrasonic tactile
  sensor integrated with tft array for force feedback and shape recognition,''
  \emph{Sensors and Actuators A: Physical}, vol. 271, pp. 348--355, 2018.

\bibitem{shinoda1997acoustic}
H.~Shinoda, K.~Matsumoto, and S.~Ando, ``Acoustic resonant tensor cell for
  tactile sensing,'' in \emph{Proceedings of International conference on
  Robotics and Automation}, vol.~4.\hskip 1em plus 0.5em minus 0.4em\relax
  IEEE, 1997, pp. 3087--3092.

\bibitem{tanaka2015tactile}
Y.~Tanaka, T.~Fukuda, M.~Fujiwara, and A.~Sano, ``Tactile sensor using acoustic
  reflection for lump detection in laparoscopic surgery,'' \emph{International
  journal of computer assisted radiology and surgery}, vol.~10, pp. 183--193,
  2015.

\bibitem{teramoto2001acoustical}
K.~Teramoto and K.~Watanabe, ``Acoustical tactile sensor utilizing multiple
  reflections for principal curvature measurement,'' in \emph{SICE 2001.
  Proceedings of the 40th SICE Annual Conference. International Session Papers
  (IEEE Cat. No. 01TH8603)}.\hskip 1em plus 0.5em minus 0.4em\relax IEEE, 2001,
  pp. 339--344.

\bibitem{park2022biomimetic}
K.~Park, H.~Yuk, M.~Yang, J.~Cho, H.~Lee, and J.~Kim, ``A biomimetic
  elastomeric robot skin using electrical impedance and acoustic tomography for
  tactile sensing,'' \emph{Science Robotics}, vol.~7, no.~67, p. eabm7187,
  2022.

\bibitem{calli2015benchmarking}
B.~Calli, A.~Walsman, A.~Singh, S.~Srinivasa, P.~Abbeel, and A.~M. Dollar,
  ``Benchmarking in manipulation research: The ycb object and model set and
  benchmarking protocols,'' \emph{arXiv preprint arXiv:1502.03143}, 2015.

\bibitem{zoller2020active}
G.~Z{\"o}ller, V.~Wall, and O.~Brock, ``Active acoustic contact sensing for
  soft pneumatic actuators,'' in \emph{2020 IEEE International Conference on
  Robotics and Automation (ICRA)}.\hskip 1em plus 0.5em minus 0.4em\relax IEEE,
  2020, pp. 7966--7972.

\bibitem{wall2022passive}
V.~Wall, G.~Z{\"o}ller, and O.~Brock, ``Passive and active acoustic sensing for
  soft pneumatic actuators,'' \emph{arXiv preprint arXiv:2208.10299}, 2022.

\bibitem{shinoda1996tactile}
H.~Shinoda and S.~Ando, ``A tactile sensor with 5-d deformation sensing
  element,'' in \emph{Proceedings of IEEE International Conference on Robotics
  and Automation}, vol.~1.\hskip 1em plus 0.5em minus 0.4em\relax IEEE, 1996,
  pp. 7--12.

\bibitem{zoller2018acoustic}
G.~Z{\"o}ller, V.~Wall, and O.~Brock, ``Acoustic sensing for soft pneumatic
  actuators,'' in \emph{2018 IEEE/RSJ International Conference on Intelligent
  Robots and Systems (IROS)}.\hskip 1em plus 0.5em minus 0.4em\relax IEEE,
  2018, pp. 6986--6991.

\bibitem{rajendran2023towards}
V.~Rajendran, S.~Parsons, and A.~Ghalamzan, ``Towards continuous acoustic
  tactile soft sensing,'' \emph{International Conference on Automation and
  Robotics}, 2023.

\bibitem{rajendran20232d}
V.~Rajendran, ``Ast2: Single and bi-layered 2-d acoustic soft tactile skin,''
  \emph{International Conference on Soft Robotics}, 2023.

\end{thebibliography}

\end{document}